\documentclass[10pt,twocolumn,letterpaper]{article}

\usepackage{cvpr}              

\usepackage{graphicx}
\usepackage{amsmath}
\usepackage{amssymb}
\usepackage{booktabs}

\usepackage{bm}
\usepackage[pagebackref,breaklinks,colorlinks]{hyperref}

\usepackage{algorithm}
\usepackage{algpseudocode}
\usepackage{amsmath}
\usepackage{hyperref}
\usepackage{marvosym}

\usepackage[capitalize]{cleveref}
\crefname{section}{Sec.}{Secs.}
\Crefname{section}{Section}{Sections}
\Crefname{table}{Table}{Tables}
\crefname{table}{Tab.}{Tabs.}
\usepackage{silence}
\makeatletter
\robustify\@latex@warning@no@line
\def\thanks#1{\protected@xdef\@thanks{\@thanks
        \protect\footnotetext{#1}}}
\makeatother
\usepackage{authblk}

\def\cvprPaperID{4127} 
\def\confName{CVPR}
\def\confYear{2022}

\begin{document}

\title{Universal Domain Adaptive Object Detector}

\author{Wenxu Shi$^1$, Lei Zhang$^{1(}$\textsuperscript{\textrm{\Letter}}$^)$, Weijie Chen$^2$, Shiliang Pu$^2$\\
$^1$Learning Intelligence \& Vision Essential (LiVE) Group\\
$^1$School of Microelectronics and Communication Engineering, Chongqing University, China\\
$^2$Hikvision Research Institute\\
{\tt\small \{wxshi,leizhang\}@cqu.edu.cn,}
{\tt\small \{chenweijie5,pushiliang.hri\}@hikvision.com}
}

\maketitle

\begin{abstract}
   Universal domain adaptive object detection (UniDAOD) is more challenging than domain adaptive object detection (DAOD) since the label space of the source domain may not be the same as that of the target and the scale of objects in the universal  scenarios can vary dramatically (i.e, \emph{category shift} and \emph{scale shift}). To this end, we propose \emph{US-DAF},  namely Universal Scale-Aware Domain Adaptive Faster RCNN with Multi-Label Learning, to reduce the negative transfer effect during training while maximizing transferability as well as discriminability in both domains under a variety of scales. Specifically, our method is implemented by two modules: 1) We facilitate the feature alignment of common classes and suppress the interference of private classes by designing a Filter Mechanism  module to overcome the negative transfer caused by category shift. 2) We fill the blank of scale-aware adaptation in object detection by introducing a new Multi-Label Scale-Aware Adapter to perform individual alignment between corresponding scale for two domains.  Experiments show that \emph{US-DAF} achieves state-of-the-art results on three scenarios (\emph{i.e}, Open-Set, Partial-Set, and Closed-Set) and yields 7.1\% and 5.9\% relative improvement on benchmark datasets Clipart1k and Watercolor in particular.

\end{abstract}

\section{Introduction}
\label{sec1}
Object detection is a fundamental problem in computer vision, aiming for precise localization and classification of objects in images.  In the past few years, numerous object detection models  \cite{ref2,ref3,ref4,ref5,ref6}  based on convolutional neural networks (CNNs) \cite{ref1} have successfully improved the performance using a large amount of labeled data. Nevertheless, applying off-the-shelf pre-trained detectors to detect objects in real-world scenarios inevitably leads to a significant performance drop due to the large domain gap including object appearance, image scale, backgrounds, illumination, viewpoints, and image quality, \emph{etc}. To meet this challenge, researchers have explored domain adaptation \cite{ref15} to transfer a detector learned from a labeled source domain to an unlabeled target domain with different scenarios, which is named domain adaptive object detection (DAOD).

\begin{figure}
  \centering
  \includegraphics[width=8.0cm]{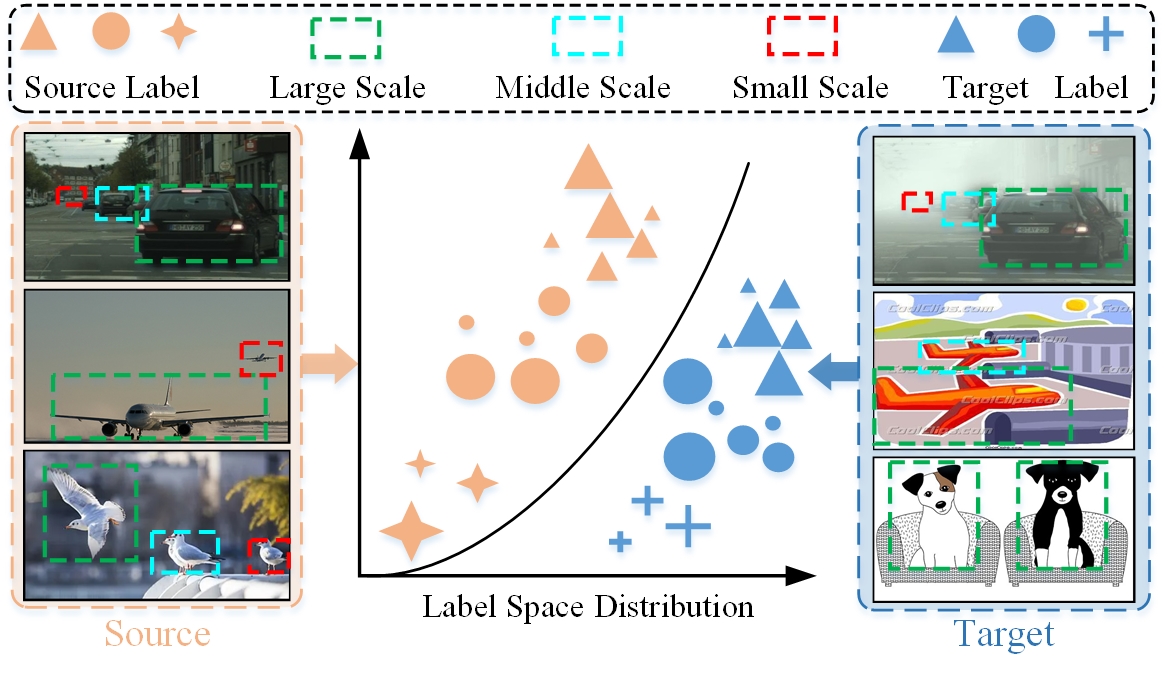}
  \caption{Problem setting of UniDAOD and illustration on the influence of scale on domain shift. 
  }\label{Figure1}

\end{figure}

Domain adaptive Faster RCNN (DAF) \cite{ref8} is the most representative DAOD work that integrates Faster R-CNN \cite{ref4} with adversarial training. To address the domain shift problem, it
aligns both the image-level and instance-level distributions across domains with adversarial training.  Subsequently, with the structural characteristics of detection tasks, DAF has rapidly evolved into a successful baseline \cite{ref9,ref10,ref11,ref12,ref13,ref14,ref43,ref44,ref45,ref46}. These methods successfully improve the performance of the detector on the target domain under the ideal and prior assumption that the label spaces are identical across domains (\emph{i.e.}, Closed-set).

Nonetheless, existing methods overlook the fact that there is  \textbf{NO} prior knowledge about the target domain categories in the real-world scenarios. Hence, as shown in Figure \ref{Figure1}, we consider a new realistic setting called Universal Domain Adaptive Object Detection (UniDAOD). For a better illustration, we denote $\mathcal{C}_s$ and $\mathcal{C}_t$  as the label set of source and target domains, respectively.  According to the relationship of label sets between the source and target domains, the universal scenarios fall into partial-set ($\mathcal{C}_s \supset  \mathcal{C}_t$), open-set ($\mathcal{C}_s \subset  \mathcal{C}_t$ or $\mathcal{C}_s \cap \mathcal{C}_t$ $\neq$ $\varnothing$, $\mathcal{C}_s \subsetneqq  \mathcal{C}_t$, $\mathcal{C}_s \supsetneqq  \mathcal{C}_t$ ), and closed-set ($\mathcal{C}_s =  \mathcal{C}_t$) scenarios.  Thus, the main task for UniDAOD is to recognize the common classes  (\emph{i.e.}, classes shared across domains)  and eliminate the domain gap, by simultaneously suppressing the interference of private classes (\emph{i.e.}, classes only exist in one domain). Besides, Figure \ref{Figure1} shows that the distributions of features extracted from objects at different scales can be very different due to the perspective projection effect, \emph{e.g.}, cars that are far away are usually very small in an image, while the near ones are relatively larger. Thus, a uniform feature alignment across all scales, as previous DAOD methods did, may not be sufficient. Instead it is more feasible to perform individual alignment on each scale between domains.

Generally, the inherent challenges come from two aspects for UniDAOD.
(1) \emph{category shift challenge:} the label set of test data may not be the same as that of training data, and the private classes may lead to negative transfer due to its absence in another domain. (2) \emph{diverse scales challenge:} previous DAOD methods mainly explored the category adaptation while ignoring a crucial challenge caused by the large variance in object scales, which is difficult but important for detection performance, especially for the UniDAOD task.

To overcome these challenges, we propose an end-to-end deep universal domain adaptation framework, \emph{US-DAF}, namely Universal Scale-Aware Domain Adaptive Faster RCNN with Multi-Label Learning. Specifically,  since adversarial alignment on features of all classes without separation might hurt its discriminability (\emph{category shift challenge}), we propose a Filter Mechanism  to suppress the private classes and preserve the common classes during the adversarial training. To fill the blank of scale-aware adaptation in cross-domain object detection (\emph{diverse scale challenge}), we introduce a new Multi-Label Scale-Aware adapter to perform individual alignment between corresponding scale for two domains (\emph{i.e.}, aligning small objects to small ones, medium objects to medium ones, and large objects to large ones).

The main contributions of this paper can be summarized as the following four-fold:
\textbf{(1)} We first introduce a more practical Universal Domain Adaptive Object Detection (UniDAOD) protocol, which is accompanied with a novel Universal Scale-Aware Domain Adaptive Faster RCNN (US-DAF) framework.
\textbf{(2)} To alleviate the impact of negative transfer caused by category shift, we propose a Filter Mechanism  to reject the private classes and preserve the common classes during adversarial training on the image-level alignment and instance-level alignment.
\textbf{(3)} To tackle the problem caused by the large variation of object scales in natural scenes, we propose a new Multi-Label Scale-Aware adapter, which can leverage the scale information for better feature alignment.
\textbf{(4)} Through ablation studies and experiments, we show that our USAF achieves state-of-the-art performance and also contributes a potential baseline under this pretty new task.


\section{Related Work and Preliminaries}
\textbf{Universal Domain Adaptation:} Existing domain adaptation methods for classification \cite{ref16,ref17,ref18,ref19,ref20} generally assume that the source and target domain share identical label space. However, in real applications, it is not practical to find a source domain having the same label space as the target domain due to the diversity of detection categories. Therefore, Cao \emph{et al}. \cite{ref21} introduce the Partial Domain Adaption problem which assumes that the target label space is a subset of the source label space, and present Partial Adversarial Domain Adaptation (PADA) by down-weighing the data of outlier source classes to  alleviates negative transfer. Busto \emph{et al}. \cite{ref22} propose the Open Set Domain Adaption scene in which there is an intersection between the source and the target domain label spaces. You \emph{et al}. \cite{ref23} propose Universal Adaptation Network (UAN), equipped with a novel criterion to quantify the transferability of each sample under the generalized Universal Domain Adaptation setting that requires no prior knowledge about the label space between domains.  Fu \emph{et al}. \cite{ref24} propose Calibrated Multiple Uncertainties (CMU) with a novel transferability measure estimated by a mixture of uncertainty quantities to align target features with source features. However, directly applying these methods to object detection yields an unsatisfactory effect. The difficulty is that the image of object detection usually contains multiple objects, thus the features of an image can have complex multi-modal structures.

\begin{figure*}
  \centering
  \includegraphics[width=0.90\linewidth]{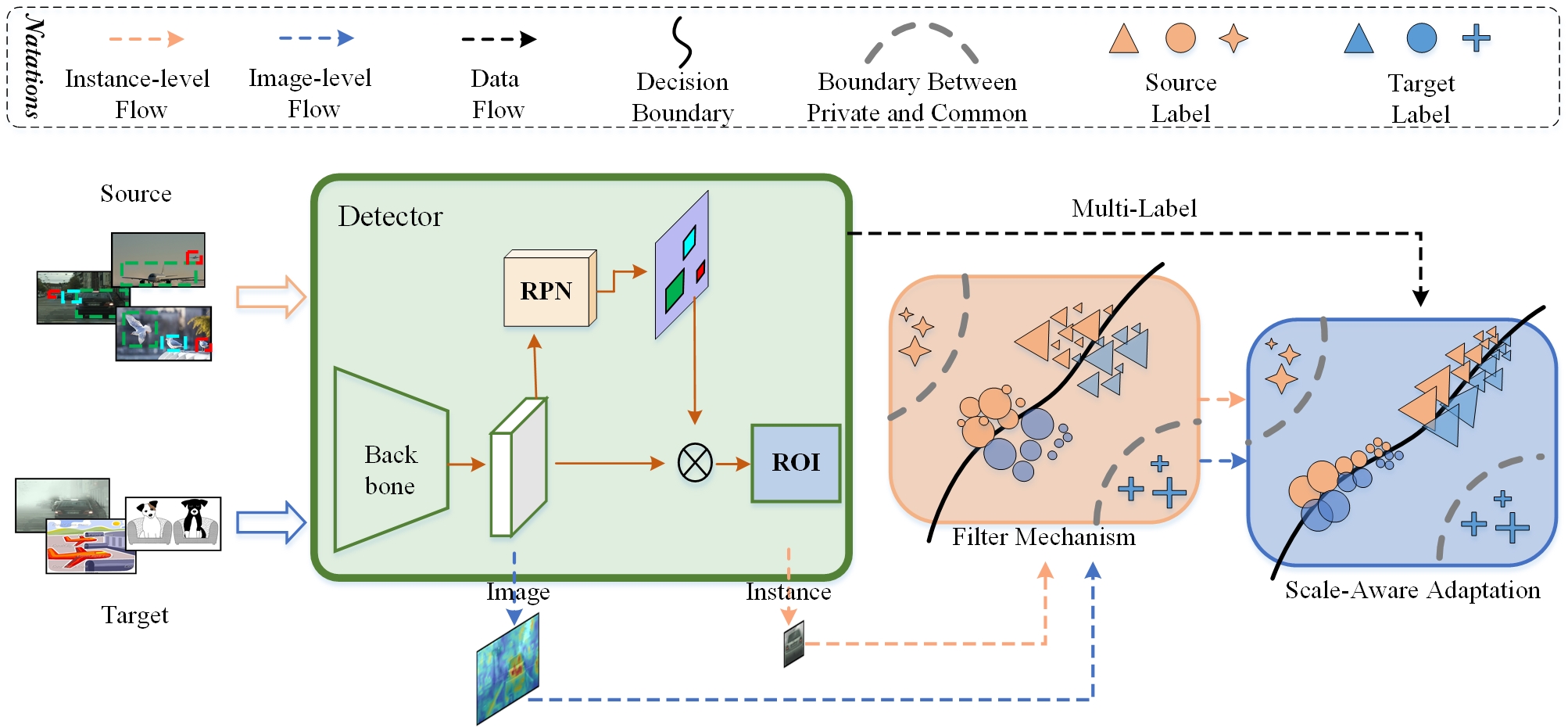}
  \caption{The architecture of the Universal Scale-Aware Domain Adaptive Faster R-CNN (US-DAF) designed for UniDAOD. Two novel modules including the Filter Mechanism (FM) and Scale-Aware Adaptation (SAA) are deployed.}\label{Figure2}
\end{figure*}

\textbf{Domain Adaptive Object Detection:} Domain adaptive object detection (DAOD) task has drawn a lot of attention due to its various applications \cite{ref8,ref9,ref10,ref11,ref12,ref13,ref14}. As a pioneering work, Chen  \emph{et al}. \cite{ref8}  propose the domain adaptive Faster-RCNN method (DAF), which achieves image-level and instance-level feature alignment by using adversarial gradient reversal. At the same time, it is pointed out that the core issue of DAOD is to solve the domain gaps in \emph{image level} and \emph{instance level}. Formally, let $d = 0$ denote that the feature is from the source domain while $d = 1$ denote that the feature is from the target domain. For the \emph{image-level} alignment,  let $D_{(u,v)}$ denote the output of the image-level domain classifier for the activation located at $(u, v)$ of the feature map, then the image-level alignment loss can be written as:
\begin{equation}\label{equ1}
  {L}_{img}=\sum\limits_{u,v}[d\log(D_{(u,v)})+(1-d)\log(1-D_{(u,v)})]
\end{equation}
For the \emph{instance-level} alignment, let $D_i$ denote the output of the instance-level domain classifier for the $i$-th region proposal, then the instance-level alignment loss is as follows:
\begin{equation}\label{equ2}
  {L}_{ins}=\sum\limits_i[d\log(D_i)+(1-d)\log(1-D_i)]
\end{equation}

After that, a large number of excellent detection algorithms emerge to overcome the image-level and instance-level domain adaption problems. Specifically, Saito \emph{et al}. \cite{ref10} utilize strong and weak domain classifiers to align local and global features. He and Zhang \cite{ref9} propose a hierarchical alignment network  that is designed to align features at different scales between the source domain and the target domain. He \emph{et al}. \cite{ref26} introduce an asymmetric tri-way approach to account for the differences in labeling statistics between domains. Chen \emph{et al}. \cite{ref27} utilize CycleGAN as a method of data augmentation to generate intermediate domain images between the source domain and the target domain to make model easy to align. Zhao \emph{et al}. \cite{ref13} use multi-label classification as an auxiliary task to regularize the features.

However,  most of the DAOD approaches have overlooked two fundamental yet practical issues: 1) All the previous methods rely on an inherent assumption that different domains have identical label space, which greatly limits their generalization in the wild. 2)  They mainly explore category adaptation and ignore the crucial challenge caused by the large variance in object scales. In this paper, we are working on solving the above two problems from two aspects: 1)  Our model considers a universal setting that imposes no prior knowledge on the label sets and proposes a filter mechanism  to suppress private classes. 2)  Our model employs a reliable  multi-label scale-aware adapter, which can leverage the scale information for better feature alignment to bridge the domain gap caused by the scale shift.

\section{METHOD}

In UniDAOD, we assume that a source domain $ \mathcal{D}_s=\{(x_i^s,y_i^s)\sim p \}_{i=1}^{n_s}$ of ${n_s}$ labeled samples from distribution $p$ and a target domain $\mathcal{D}_t=\{(x_i^t)\sim q\}_{i=1}^{n_t}$ of ${n_t}$ unlabeled  samples from distribution $q$ are provided at training. Since the label set may not be identical, we use $\mathcal{C}_s$, $\mathcal{C}_t$ to denote the label set of source and target domains, respectively.  $\mathcal{C}=\mathcal{C}_s \cap \mathcal{C}_t$ is the common label set shared by both domains, while $\overline{\mathcal{C}}_s=\mathcal{C}_s \verb|\|\mathcal{C}$ and $\overline{\mathcal{C}}_t=\mathcal{C}_t \verb|\|\mathcal{C}$ are the private label sets for source and target respectively. Note that the target label set is not accessible at training and only used for defining the UniDAOD problem. The Jaccard index of the label sets of the two domains, $\xi=\frac{\mathcal{C}_s \cap \mathcal{C}_t}{\mathcal{C}_s \cup \mathcal{C}_t}$,  is used to represent  the overlap among classes. 

\subsection{Network Structure}
To deal with the challenge (i.e., \emph{category shift} and \emph{scale shift}) mentioned in Section \ref{sec1}, we propose the Universal Scale-Aware Domain Adaptive Faster RCNN with Multi-Label Learning (US-DAF) framework, which has two steps: (1) suppresses the private classes and preserves the common classes at the image level and the instance level. (2) designs the multi-label scale-aware adapter at the image level and instance level to tackle the problem brought by the variation of object scales in natural scenes.

Since the private-class features might lead to negative transfer during the training  to hurt the discriminability of the detector, we \emph{filter out} the private classes and focus on the common classes in adapting an object detector by introducing a Filter Mechanism. To fill the blank of scale-aware adaptation in cross-domain object detection, we need to perform adaptation on the bounding box scale. The overall structure of the proposed US-DAF is presented in Figure \ref{Figure2}, and Section \ref{sec3.2} and  \ref{sec3.3} will introduce the design of the filter mechanism and the scale-aware adapter in details.
\subsection{Filter Mechanism }
\label{sec3.2}

An ideal solution for the category shift of UniDAOD is to make the samples with common categories go through for further adaptation while suppressing the samples of private categories.
If we naively pick any of the existing DAOD methods to solve the UniDAOD by aligning the source with the target domain, the private classes will impose negative transfer and degrade the detection performance of common classes in the target domain. Therefore, we adopt a sample-level Filter Mechanism. For both source and target domains, the samples with the common categories are expected to become well-aligned while the samples of private categories are expected to be ignored. Consequently, we need a criterion to explore the common category set and private category set, and then perform the adversarial domain alignment with this criterion.

Our motivation is from the observation on the optimization process with Gradient Reverse Layer (GRL) \cite{ref17}. Specifically, the objective of domain discriminator $D$ is to predict samples from source domain as 0 and samples from
target domain as 1. The ideal convergence point of the domain adversarial training is that the samples with similar categories cannot be easily distinguished, which means the predictions from domain discriminator on these samples are around the middle point 0.5. Thus, $D$ can be seen as the quantification for the domain similarity of each sample.  For a source
sample $x$, larger $D(x)$ means that it is more similar to the target domain; for a target sample $x$, smaller $D(x)$ means that it is more similar to the source domain. Therefore, we can hypothesize that $\Bbb{E}_{x\sim p_{\overline{\mathcal{C}}_s}}D(x)$ \textless $\Bbb{E}_{x\sim p_{{\mathcal{C}}}}D(x)$ \textless $\Bbb{E}_{x\sim q_{{\mathcal{C}}}}D(x)$ \textless $\Bbb{E}_{x\sim q_{\overline{\mathcal{C}}_t}}D(x)$.

\begin{figure}
  \centering
  \includegraphics[width=8.0cm]{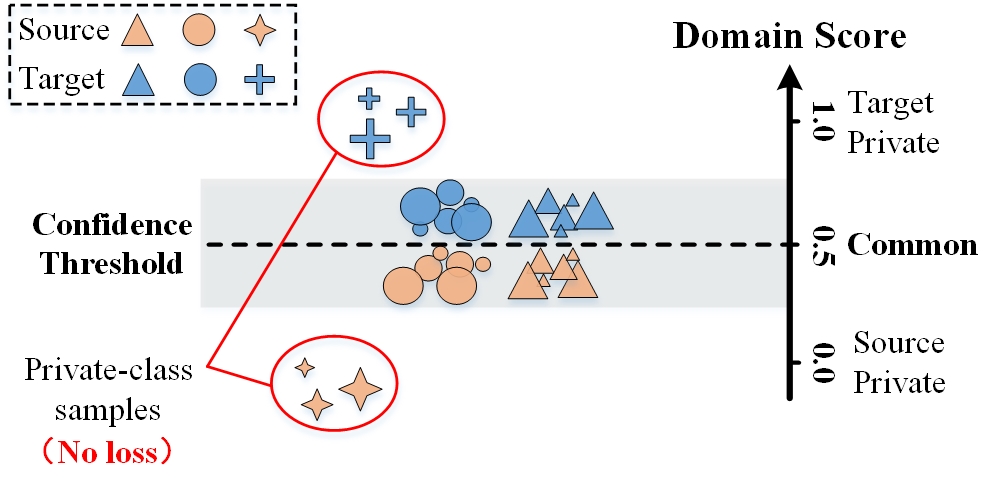}
  \caption{An overview of the filter mechanism. Samples in the bottom part are more likely to be source private class samples, while the top samples are more likely to be target private samples. We introduce a confidence threshold that allows us to  explore the common-class samples and private-class samples. }\label{Figure3}
\end{figure}

Inspired by this, we propose to draw a boundary between \emph{common} and \emph{private} points using the predictions of the domain discriminator. We visually introduce the idea in Figure \ref{Figure3}.  Specifically, the distance between the prediction and middle point, 0.5, is defined as $|D(x)- 0.5|$, where $D(x)$ is the classification output for a sample $x$. We expect that the prediction of common-class samples is closer to the  middle point than  the private-class ones. Therefore, we propose to introduce a confidence threshold parameter $m$ to explore the common category set and private category set. The above formulation shows that common-class and private-class samples can be separated with the confidence threshold parameter $m$. Note that tuning the parameter $m$ for each adaptation setting requires a validation set.

With the above analysis, by combining Filter Mechanism with image-level and instance-level alignments, the sample-level transferability criterion for the image-level domain adaptation (\emph{i.e.}, Eq. \ref{equ1}) and the instance-level domain adaptation (\emph{i.e.}, Eq. \ref{equ2}) can be respectively re-formulated as Eq. \ref{equ3} and Eq. \ref{equ4}:

\begin{equation}\label{equ3}
	{L}_{img\_FM}=\left\{
		\begin{aligned}
		 {L}_{img} &, ~~~~~ if~~ (\left| D_{(u,v)}- 0.5 \right| \textless m)  \\
		~0~~~~&, ~~~~~otherwise
	    \end{aligned}
	\right.
\end{equation}

\begin{equation}\label{equ4}
	{L}_{ins\_FM}=\left\{
		\begin{aligned}
		 {L}_{ins} &, ~~~~~if~~ (\left| D_{i}- 0.5 \right| \textless m)  \\
		~0~~~~&, ~~~~~otherwise
	    \end{aligned}
	\right.
\end{equation}

The introduction of the confidence threshold $m$ allows us to give the final separation loss for differentiating the common-class samples from private-class samples.

\subsection{Scale-Aware Adaptation with Multi-Label Learning}
\label{sec3.3}

\begin{figure}
  \centering
  \includegraphics[width=8.0cm]{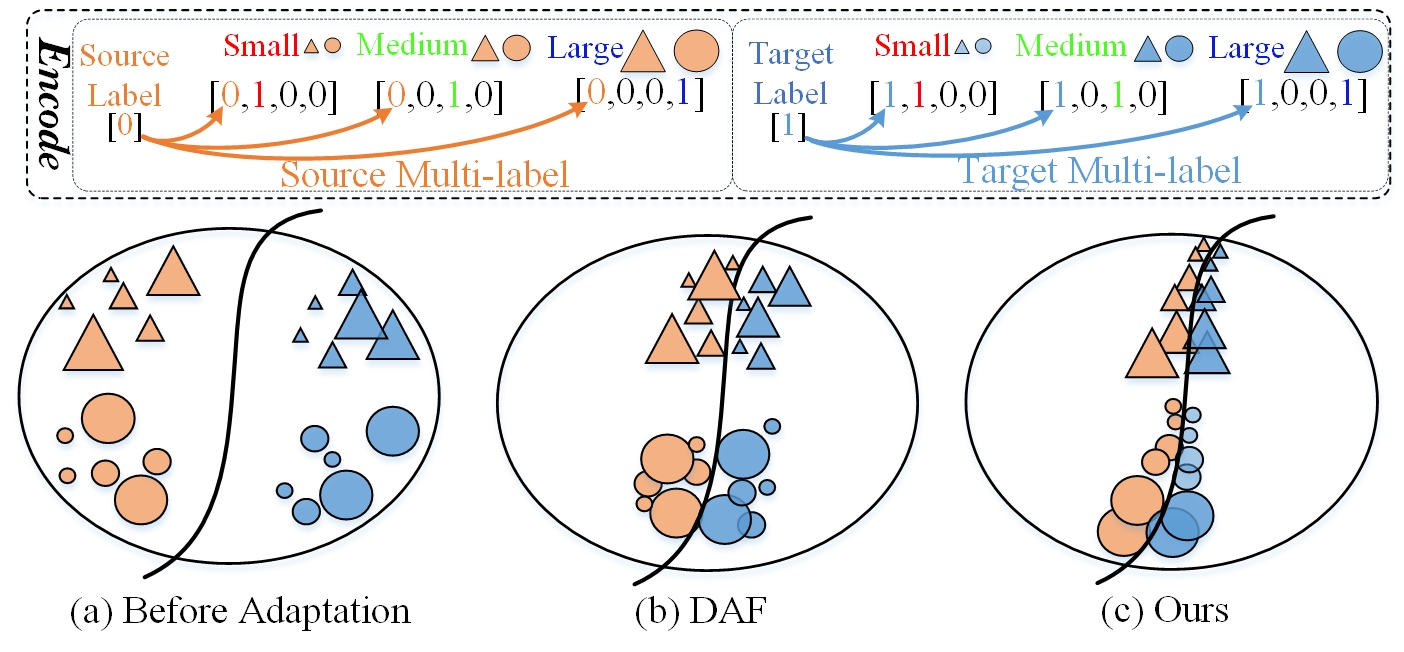}
  \caption{Illustration on different strategies for feature alignment. Although the domain gap is reduced successfully by DAF \cite{ref8}, there might exist misaligned the scales between two domains. Our scale-aware adaptation  with multi-label learning is expected to avoid such misalignment.}\label{Figure4}
\end{figure}

We design a scale-aware adaptation (SAA) module to leverage the scale information for better feature alignment at image level and instance level. Our motivation lies in two aspects. First, Chen \cite{ref32} and Lin \cite{ref33} claim that the scale of objects in natural images can vary dramatically, which is an inevitable and non-negligible problem in image segmentation. We can therefore make a reasonable assumption that the large variance in object scales often brings a crucial challenge to cross-domain object detection. Second,  as discussed in Section \ref{sec1}, current DAOD models \cite{ref8,ref9,ref10,ref26,ref27,ref13} ignore the importance of scale-aware alignment and have a uniform feature alignment across all scales, which may not be sufficient (see Figure \ref{Figure4} for illustration).

Building upon those above considerations, we introduce the scale-aware adaptation module to perform alignment between the corresponding scale for two domains.  Intuitively, the size of instance features can be divided into three categories: small ($\textless 20^2$ pixels), medium ($20^2 \sim 100^2$ pixels), and large ($\textgreater 100^2$ pixels). The size of the instance is cost-free for detection datasets, and can be easily acquired through the sub-module RPN of Faster RCNN. It is worth noting that, as illustrated in Figure \ref{Figure2}, we attach the scale-aware adaptation with the instance level and image level, due to the image level features contain fine-grained information associated with the objects in the instance level.

In particular, we treat the scale-aware domain classification task as a multi-label classification problem \cite{ref34,ref35}.  It takes the size of the instance  features produced by the RPN of the Faster-RCNN model as additional label input, and combines it with the original domain label. More formally, as shown in Figure \ref{Figure4}, we define by $d$ the multi-label of a training image, and encode $d=[0,1,0,0]$, $d=[0,0,1,0]$, and $d=[0,0,0,1]$ for the source domain with three different scales and $d=[1,1,0,0]$, $d=[1,0,1,0]$, and $d=[1,0,0,1]$ for the target domain with three different scales. It is worth noting that the first  entry of the encoded multi-label $d$ is used to indicate the domain and the last three entries indicate the scale (small, medium, large) of objects.  The domain-invariant features
can then be learned by minimizing the following  multi-label  cross-entropy loss:
 \begin{equation}\label{equ5}
  {L}^m_{img}=\sum\limits_{u,v}[d^T\log(D^m_{(u,v)})+(1-d)^T\log(1-D^m_{(u,v)})]
\end{equation}

\begin{equation}\label{equ6}
  {L}^m_{ins}=\sum\limits_i[d^T\log(D^m_i)+(1-d)^T\log(1-D^m_i)]
\end{equation}

 With the above analysis, by combining the filter mechanism(\emph{i.e.}, Eq.\ref{equ3} and Eq.\ref{equ4}) with SAA , the sample-level transferability criterion for the image-level multi-scale domain adaptation (\emph{i.e.}, Eq.\ref{equ5}) and the instance-level multi-scale domain adaptation (\emph{i.e.},  Eq.\ref{equ6}) can be respectively defined as Eq.\ref{equ7} and Eq.\ref{equ8}:
 \begin{equation}\label{equ7}
	{L}^m_{img\_FM}=\left\{
		\begin{aligned}
		 {L}^m_{img} &, ~~~~~ if~~ (\left| D^m_{(u,v)}[0]- 0.5 \right| \textless m)  \\
		~0~~~~&, ~~~~~otherwise
	    \end{aligned}
	\right.
\end{equation}

\begin{equation}\label{equ8}
	{L}^m_{ins\_FM}=\left\{
		\begin{aligned}
		 {L}^m_{ins} &, ~~~~~if~~ (\left| D^m_{i}[0]- 0.5 \right| \textless m)  \\
		~0~~~~&, ~~~~~otherwise
	    \end{aligned}
	\right.
\end{equation}
where $D^m_{(u,v)}[0]$ and $D^m_{i}[0]$ represent the first entry of the prediction output vectors $D^m_{(u,v)}$ and $D^m_{i}$ generated by the multi-label multi-scale domain discriminator, indicating the probability of the presence of objects from the source or target.
\subsection{ Overall End-to-End Learning}

The overall framework of US-DAF with a detailed pipeline can be observed in Figure \ref{Figure2}. US-DAF contains three loss functions, including the detection
loss ${L}_{DET}$, image-level domain adversarial loss ${L}^m_{img\_FM}$ and instance-level domain adversarial loss ${L}^m_{ins\_FM}$. The standard detection loss ${L}_{DET}$
in Faster-RCNN \cite{ref4} is used, \emph{i.e.}, the cross-entropy loss is used for classification and the SmoothL1 loss is used for regression (localization). Note that the detection loss is only
optimized on the labeled source samples.

The combination of the last two losses formulates the proposed universal domain alignment (UniDA) of \emph{US-DAF} in both image-level and instance-level. By jointly considering
Eq.\ref{equ7} and Eq.\ref{equ8}, the proposed UniDA loss is expressed as:
\begin{equation}\label{equ9}
  {L}_{UniDA}={L}^m_{img\_FM}+ {L}^m_{ins\_FM}
\end{equation}
With the combination of the detection loss and domain alignment loss, the final loss of the proposed \emph{US-DAF} can be written as:
\begin{equation}\label{equ10}
  {L}_{US-DAF}=\min\limits_{G}\max\limits_{D}  {L}_{DET}- \eta {L}_{UniDA}
\end{equation}
where $\eta$ is a hyper-parameter, \emph{G} denotes a Faster R-CNN object detector, and \emph{D} indicates the domain classifier. The
mini-max adversarial optimization is implemented by the GRL \cite{ref17}.
\begin{table*}[t]
\centering
\caption{Results (\%) on universal adaptation from {\bf Pascal VOC \cite{ref7} to Clipart1k \cite{ref40}} (Open-set, $\xi$=0.75). The source private classes include: \emph{`train' and `tvmonitor'}. The target private classes include: \emph{`aeroplane', `bicycle' and `bird'}. US-DAF$^\dagger$ denotes the ablation analysis without filter mechanism. US-DAF* means that the ablation analysis without scale-aware adaptation.\label{Table1}}
   \setlength{\tabcolsep}{1.2mm}
   \begin{tabular}{l|ccccccccccccccc|c}
      \toprule
      Methods          & boat & bottle    & bus     & car    & cat    & chair    & cow &table &dog  &horse &motor &person &plant &sheep& sofa     & mAP\\
      \midrule
      \midrule
      Faster-RCNN         &31.8   &41.2      &31.1   &34.7    &5.1   &33.7      &23.0     &20.7  &8.3  &43.0   &52.7      &49.6   &40.6    &17.0   &13.8      &29.8   \\
      DAF\cite{ref8}      &\textbf {37.2}   &38.0      &26.9   &35.9    &2.3   &35.2      &24.0     &28.5  &4.2  &33.8   &54.7      &59.4   &\textbf {58.4} &13.4  &17.9   &31.3 \\
      MAF\cite{ref9}      &24.2   &42.9      &35.1   &32.3    &11.0   &\textbf {41.7}     &22.4     &\textbf {32.6}  &6.7  &40.0   &59.1      &52.7   &41.0    &24.1   &17.9     &32.2 \\
      HTCN\cite{ref27}    &25.9   &\textbf {47.8}      &36.0   &32.8    &11.3   &39.4     &51.7 &18.7  &\textbf {10.5}  &40.9   &56.3      &57.9  &49.4    &21.3   &20.4      &34.7 \\
      UAN\cite{ref23}     &26.6   &37.7      &48.2   &31.5    &8.6    & 32.8    &23.7     &31.6  &2.4   &36.6   &56.6      &42.8   &44.8    &14.7  &16.4     &30.3 \\
      CMU\cite{ref24}    &14.7  &41.9      &\textbf {52.5}   &34.7   &9.2    & 36.5    &38.1    &21.0  &7.6   &37.0  &48.6      &55.7   &44.5    &17.7  &21.1     &32.1 \\
      \midrule
      \midrule
      US-DAF$^\dagger$   &32.4   &38.6  &30.8  &\textbf {39.0}    &13.2  &37.2     &61.5     &29.7  &8.2   &45.1   &63.2      &61.5  &43.7      &\textbf {28.4}    &22.8   &37.0\\
      US-DAF*  &33.5   &45.9      &27.9   &35.1    &15.1  &39.2      &53.1     &25.9  &9.8   &42.3   &\textbf {65.3}      &61.6   &48.4    &25.0   &17.4 &36.4\\
      US-DAF(ours)      &34.9   &40.8     &28.9  &36.4     &\textbf {17.7}    &38.4     &\textbf {64.6}    &28.0  &10.3   &\textbf {45.8}   &64.5      &\textbf {62.5}   &52.1    &25.8   &\textbf {24.8}    &\textbf {38.4}\\
      \bottomrule
   \end{tabular}

\end{table*}

\begin{table*}[t]
\centering
\caption{Results (\%) on universal adaptation from {\bf Pascal VOC  to Clipart1k} (Open-set, $\xi$=0.5). The source private classes include: \emph{`aeroplane', `bicycle', `bird', `boat', and `bottle'}. The target private classes include:  \emph{`plant', `sheep', `sofa', `train', and 'tvmonitor'}.  \label{Table2}}
   \setlength{\tabcolsep}{3.1mm}
   \begin{tabular}{l|cccccccccc|c}
      \toprule
      Methods           & bus     & car    & cat    & chair    & cow &table &dog  &horse &motor &person    & mAP\\
      \midrule
      \midrule
      Faster-RCNN           & 43.3   &33.0    &8.4    &32.1      &24.0    &28.7  &6.9  &34.9   &51.8      &42.5        &30.6   \\
      DAF\cite{ref8}        & 37.5   &32.8    &10.2   &40.3      &27.2    &\textbf {31.3}  &4.1  &41.0   &55.5      &52.0        &33.2 \\
      MAF\cite{ref9}        & 37.1   &31.1    &9.7    &38.1      &19.9    &29.1  &2.5  &37.3   &50.7      &50.0        &30.6 \\
      HTCN\cite{ref27}      & 29.5  &34.4    &\textbf {17.3}   &33.8      &50.6    &14.0  &3.6 &\textbf {46.9}   &\textbf {74.7}     &58.5         &36.3 \\
      UAN\cite{ref23}       &\textbf {48.9}    &26.4    &14.6   &36.7      &49.9    &30.0  &3.2  &39.9   &56.1      &52.0        &35.8 \\
        CMU\cite{ref24}       &33.3    &32.8    &8.1   &41.5      &55.5    &24.6  &5.6  &43.3   &54.9      &60.4       &36.0 \\
      \midrule
      \midrule
      US-DAF$^\dagger$     &36.1    &33.3    &11.8   &39.9      &58.2   &26.1  &7.5  &43.2  &70.5      &57.8         &38.4\\
      US-DAF*      &34.8    &39.8    & 11.9   &37.4      &55.2    &27.4  &\textbf {16.8}  &34.5  &59.8      &\textbf {64.1}         &38.2\\
     US-DAF(ours)               &31.3   &\textbf {41.9}    &7.3   &\textbf {42.9}      &\textbf {64.3}    &30.0  &5.7  &44.8  &69.5      &61.9         &\textbf {40.0}\\
      \bottomrule
   \end{tabular}

\end{table*}

\begin{table}[t]
\centering
\caption{Results (\%) on universal adaptation from {\bf Pascal VOC to Clipart1k} (Open-set, $\xi$=0.25). The source private classes include:  \emph{ `bus', `car', `cat', `chair', `cow', `table', and `dog'}. The target private classes include:  \emph{`horse', `motorbike', `person', `plant', `sheep', `sofa', `train', and `tvmonitor'}.  \label{Table3}}
   \setlength{\tabcolsep}{1.3mm}
   \begin{tabular}{l|ccccc|c}
      \toprule
      Methods           & plane     & bicycle    & bird    & boat    & bottle     & mAP\\
      \midrule
      \midrule
      Faster-RCNN           & 33.2   &55.7    &25.4   &29.2      &41.6            &37.0   \\
      DAF\cite{ref8}        & 31.5   &42.5    &25.2   &\textbf {34.4}      &\textbf {50.8}          &36.9 \\
      MAF\cite{ref9}       & 29.3   &57.0    &27.1   &33.9     &41.8          &37.8 \\
      HTCN\cite{ref27}      & 32.5  &53.0    &24.1  & 27.0     &48.4           &37.0 \\
      UAN\cite{ref23}      & 35.6   &55.9    &27.1   &28.2      &44.2           &38.2 \\
      CMU\cite{ref24}       &\textbf {45.5}   &52.7    &\textbf {28.8}   &29.4     &40.1   &39.3 \\
      \midrule
      \midrule
      US-DAF$^\dagger$       & 43.2   &54.2    &24.1   &28.6     &43.6           &38.7\\
      US-DAF*       & 41.9   &54.7    &25.4   &26.0      &41.7           &37.9\\
      US-DAF(ours)          & 44.2  &\textbf {57.5}   &27.9   &32.2      &40.5           &\textbf {40.5}\\
      \bottomrule
   \end{tabular}
\end{table}
\section{Experiments}
To perform a thorough evaluation  under a variety of UniDAOD settings, we compare US-DAF with state of the art methods tailored to DAOD settings on several datasets with different scenarios, \emph{i.e.}, {\bf open-set}, {\bf partial-set}, and {\bf closed-set}. We conduct sufficient experiments and evaluate our proposed method on benchmark datasets, including Cityscapes \cite{ref36}, Foggy Cityscapes \cite{ref37}, PASCAL VOC \cite{ref7}, Clipart1k and WaterColor \cite{ref40}. Then, we explore the performance with respect to the change of $\xi$.  Code will be available.

\subsection{Experimental Setup}

{\bf Implementation Details.} For fair comparison, the backbone network of our proposed US-DAF model is ResNet101 \cite{ref41} pre-trained on ImageNet \cite{ref1} in the experiments. Following the default settings in \cite{ref8}, the shorter side of each input image is resized to 600 pixels. We optimize the network by using the stochastic gradient descent (SGD) optimizer with a momentum of 0.9 and a weight decay of 0.0005. The initial learning rate is set to 0.001 and dropped to 0.0001 after 50k iterations. Totally, 100k iterations are trained. The trade-off parameter $\eta$ in Eq.\ref{equ10} is set as 0.01 in our implementation. A single batch is composed of two images respectively for the source and target domains. To evaluate the adaptation performance, we report mean average precision (mAP) with IOU threshold of 0.5.

{\bf Compared Methods.} We compare the proposed US-DAF with (\textbf{1}) \emph{CNN based object detection}: Source only {\bf Faster RCNN} \cite{ref4} without any adaption, (\textbf{2}) \emph{Traditional domain adaptive object detection methods}: Domain Adaptive Faster R-CNN ({\bf DAF}) \cite{ref8}, Multi-adversarial Faster-RCNN ({\bf MAF}) \cite{ref9}, and  Hierarchical Transferability Calibration Network ({\bf HTCN}) \cite{ref27}, (\textbf{3}) \emph{Partial domain adaptation methods}: Partial Adversarial Domain Adaptation ({\bf PADA}) \cite{ref21}, (\textbf{4}) \emph{Universal domain adaptation methods}: Universal Adaptation Network ({\bf UAN}) \cite{ref23},  Calibrated Multiple Uncertainties ({\bf CMU}) \cite{ref24}. Because these methods achieved state-of-the-art performance in their respective task, it is valuable to show their performance in the UniDAOD setting. It is worth noting that PADA, UAN, and CMU are domain adaptation methods for image classification, and we use them in domain adaptation object detection.

\subsection{Experimental Results}
In the case of different $\xi$, with respect to open-set, partial-set, and closed-set, the mean average precision of the common classes are shown in Tables \ref{Table1} to \ref{Table6}. US-DAF outperforms all the compared methods in terms of the mean average precision. These consistent results suggest that US-DAF can overcome the double challenge brought by \emph{category shift}  and \emph{scale issue} between the source and target domains. We have the following observations.

{\bf Open-set scenario.}  As shown in Tables \ref{Table1} to \ref{Table3}, we use the PASCAL VOC \cite{ref7} as the source domain and the Clipart1k \cite{ref40} as the target domain, and we select some classes as the common classes or private classes. Specifically, we design three experiments for this scenario (from PASCAL VOC to Clipart1k) with different $\xi=\{0.75, 0.5, 0.25\}$ .

The experiments show that whatever this $\xi$ is, our US-DAF can achieve state-of-the-art results among all compared methods.  The proposed US-DAF clearly
outperforms the baseline model DAF \cite{ref8} by +7.1\%, +6.8\%, and +3.6\% with different  $\xi$. Note that our US-DAF also can surpass the MAF \cite{ref9} and HTCN \cite{ref27}, even if they have a multi-layer alignment structure and additional adaptation modules. And both source and target domains have their own private classes in these scenarios, which lead to more serious negative transfer. However, our model still performs well in these scenarios by using the proposed filter mechanism and scale-aware adaptation.

Furthermore, in open-set settings, especially the difficult task Watercolor \cite{ref30} $\rightarrow$ PASCAL VOC \cite{ref7} (\emph{i.e.}, Table \ref{Table4}),  most existing methods perform similarly to or even worse than Faster RCNN, indicating that existing methods are prone to negative transfer in open-set settings. That is, they perform worse than a model only trained on source data without any adaptation. We can find that DAF \cite{ref8}, MAF \cite{ref9}, and HTCN \cite{ref27} suffer from negative transfer in most classes and are only able to promote the adaptation for a few classes. Comparatively, the proposed US-DAF promotes positive transfer for all classes.

{\bf Partial-set scenario.}  We conduct the partial domain adaptive object detection scenario, in which the target label set is completely a subset of the source label set ($\mathcal{C}_s \supset  \mathcal{C}_t$). WaterColor \cite{ref40} dataset contains 6 categories in common with PASCAL VOC \cite{ref7}. Therefore, we adopt the PASCAL VOC as the source domain and the WaterColor as the target domain in the partial domain adaptation.

The results are presented in Table \ref{Table5}. We can see that our US-DAF achieves 55.2\% mAP, which outperforms a remarkable increase of +5.9\% over the baseline DAF \cite{ref8}. Furthermore, we can observe that most existing DAOD methods perform similarly to or even worse than Faster RCNN, indicating that existing methods are prone to negative transfer in partial-set settings. That is, they perform worse than a model only trained on source data without any adaptation. Note that our US-DAF also can surpass the UAN \cite{ref23} and CMU \cite{ref24}, even if they avoid negative transfer in most tasks. Comparatively, the proposed US-DAF promotes positive transfer for all classes. These consistent results suggest that US-DAF can overcome the double challenge brought by  category shift and scale issue between the source and target domains.

\begin{table}[t]
\centering
\caption{Results (\%) on universal adaptation from {\bf  Watercolor \cite{ref30} to Pascal VOC \cite{ref7}}. (Open-set $\mathcal{C}_s \subset  \mathcal{C}_t$). \label{Table4}}
   \setlength{\tabcolsep}{0.7mm}
   \begin{tabular}{l|cccccc|c}
      \toprule
      Methods           & bicycle     & bird    & car    & cat    & dog  & person   & mAP\\
      \midrule
      \midrule
      Faster-RCNN           & 29.8   &50.2      &47.1   &62.2      &51.5         &57.8        &49.8   \\
      DAF\cite{ref8}        & 29.5   &  \textbf {53.8}    & 50.6   & 58.1      &  48.1     &  56.5     &  49.4 \\
      MAF\cite{ref9}        & 28.5   &50.0    &46.8   &59.4      &50.2     &58.6     &48.9 \\
      HTCN\cite{ref27}       & 26.4    &43.0    &46.5  &50.8     &44.0     &53.9      & 44.1 \\
      PADA\cite{ref21}      & 32.5    &52.2      &51.8    &57.7       &54.2        &\textbf {60.2}        &51.4 \\
      UAN\cite{ref23}       & 33.6   &52.1      &\textbf {53.8}    &62.4       &52.2        &56.1        &51.7 \\
       CMU\cite{ref24}  &\textbf {36.9}   &51.2      &53.3   &59.3      &51.7        &59.9        &52.0 \\
      \midrule
      \midrule
      US-DAF$^\dagger$       & 28.7  &53.1   &51.3   &62.4    &53.4      &58.7    &  51.3\\
      US-DAF*                &30.6   &  52.4    & 54.5   & 61.4      &  53.8      &  59.9      &  52.1\\
      US-DAF(ours)          & 35.0   &  52.4   &52.7   & \textbf {63.1}     &  \textbf {54.3}      & 59.8      & \textbf { 52.9}\\

      \bottomrule

   \end{tabular}

\end{table}

\begin{table}[t]
\centering
\caption{Results (\%) on universal adaptation from {\bf Pascal VOC to Watercolor }  in partial scenario. (Partial-set $\mathcal{C}_s \supset  \mathcal{C}_t$).  \label{Table5}}
   \setlength{\tabcolsep}{0.7mm}
   \begin{tabular}{l|cccccc|c}
      \toprule
      Methods           & bicycle     & bird    & car    & cat    & dog  & person   & mAP\\
      \midrule
      \midrule
      Faster-RCNN           & 82.4   &51.7    &48.4   &39.9      &30.7    &59.2        &52.0   \\
      DAF\cite{ref8}        & 73.4   &51.9    &43.1   &35.6      &28.8    &63.1        &49.3 \\
      MAF\cite{ref9}        & 70.4   &50.3    &44.3   &36.7      &30.6    &62.9        &49.2 \\
      HTCN\cite{ref27}      & 74.1  &49.8    &\textbf {51.9}   &35.3      &35.3    &66.0        &52.1 \\
      PADA\cite{ref21}      & 74.3   &53.6    &45.6   & 38.6      & \textbf {41.8}    &64.3        &53.0 \\
      UAN\cite{ref23}       & 78.0   &53.6    &50.4   &36.4      &35.8    &65.6        &53.3 \\
       CMU\cite{ref24}       & 82.0   &53.9    &48.6   &39.6      &33.1    &\textbf {66.0}        &53.9 \\
      \midrule
      \midrule
      US-DAF$^\dagger$      & 81.3   &52.7   &51.5   &37.5      &35.9    &61.9        &53.5 \\
      US-DAF*       & 81.8   & \textbf {55.3}  & 44.5 & 38.8    &31.1     &62.1      &  52.3\\
     US-DAF(ours)          & \textbf {86.5}   &  54.1  &  50.0  & \textbf {43.0}   &34.0     &  63.2     &   \textbf {55.2}\\
      \bottomrule
   \end{tabular}
\end{table}

\begin{table}[t]
\centering
\caption{Results (\%) on domain adaptation from {\bf  Cityscape \cite{ref36} to Foggy Cityscape \cite{ref37}}  in
closed-set scenario. (closed-set $\mathcal{C}_s = \mathcal{C}_t$). ``+FM"  and ``+SAA" mean the replacement of the original mechanism with the proposed filter mechanism and scale-aware adaptation module, respectively.\label{Table6}}

   \setlength{\tabcolsep}{0.25mm}
   \begin{tabular}{l|cccccccc|c}
      \toprule
      Methods              & prsn  & rider    & car    & trunk    &bus   &train  & mcyc  & bicy   & mAP\\
      \midrule
      \midrule
      Faster-RCNN               & 17.8   &23.6     &27.1   &11.9     &23.8    &9.1    &14.4  &22.8    &18.8\\
      \midrule
      \midrule
      DAF\cite{ref8}        & 25.0  &31.0      &40.5   &22.1    &35.3    &20.2   &20.0  &27.1    &27.6\\
      DAF+FM                 & 27.4 &39.6      &42.0             &22.3       &35.0             &11.8  &20.2    &32.8&28.7\\
      DAF+SAA                 & \textbf {31.8}  &\textbf {44.2}      &\textbf {44.4}   &\textbf {22.8}    &\textbf {38.7}    &\textbf {31.1}    &\textbf {28.0}  &\textbf {35.7}    &\textbf {34.6}\\
      \midrule
      \midrule
       MAF\cite{ref9}        & 28.2   &39.5    &43.9   &23.8      &39.9   &\textbf {33.3}     &29.2  &33.9    &34.0\\
       MAF+FM                 &33.4   &44.6    &44.6   &23.9      &37.8            &28.8         &30.2    &\textbf {37.3}  &35.1\\
       MAF+SAA                 &\textbf { 33.9 }  &\textbf {47.0}    &\textbf {50.6}   &\textbf {28.1}      &\textbf {46.7}   &27.7     &\textbf {32.2} &36.4    &\textbf {37.8}\\
      \bottomrule
   \end{tabular}
\end{table}

{\bf Closed-set scenario.}  Existing DAOD methods work under the closed-set domain adaptation setting, where the category sets of the source and target domains are the same. Therefore, we use the samples from common label set to compare our methods with previous methods. As shown in Table \ref{Table6}, we conduct the experiment on the closed set ($\mathcal{C}_s = \mathcal{C}_t$) from Cityscapes \cite{ref36} to Foggy Cityscapes \cite{ref37} by comparing the two baseline methods \cite{ref8,ref9}.

Experimental result shows that our proposed US-DAF outperforms the two methods, which significantly indicates that our the sample-level transferability criterion filter mechanism of US-DAF does not deteriorate performance on the closed set domain adaptation setting, and demonstrates the effectiveness of our scale-aware adaptation approach on the closed-set domain adaptation scenario.

\subsection{Further Empirical Analysis}
In this section, we conduct model analysis and discussion to investigate the effect of our US-DAF for the UniDAOD task. An in-depth insight into the proposed models is shown.


{\bf Ablation Study.}
We conduct the ablation study to show the effectiveness of each component (\emph{i.e.}, FM, SAA) by evaluating several variants of US-DAF and the results are reported at the bottom part of Tables \ref{Table1} to \ref{Table5}  in all scenarios. We can see that the proposed filter mechanism (FM) is designed reasonably and when it is removed, the performance drops accordingly. Take Pascal VOC $\rightarrow$ Clipart1k ($\xi=0.5$) (\emph{i.e.}, Table \ref{Table2}) as an example, with FM, its mean average precision is 40.0\%, however, if without FM, its accuracy drops to 38.4\%. Similarly, the results from Tables \ref{Table1} to \ref{Table5} also show that removing the SAA can make the performance correspondingly degrade. This indicates that the SAA module in the US-DAF is designed reasonably. 
\begin{figure}
  \centering
  \includegraphics[width=0.95\linewidth]{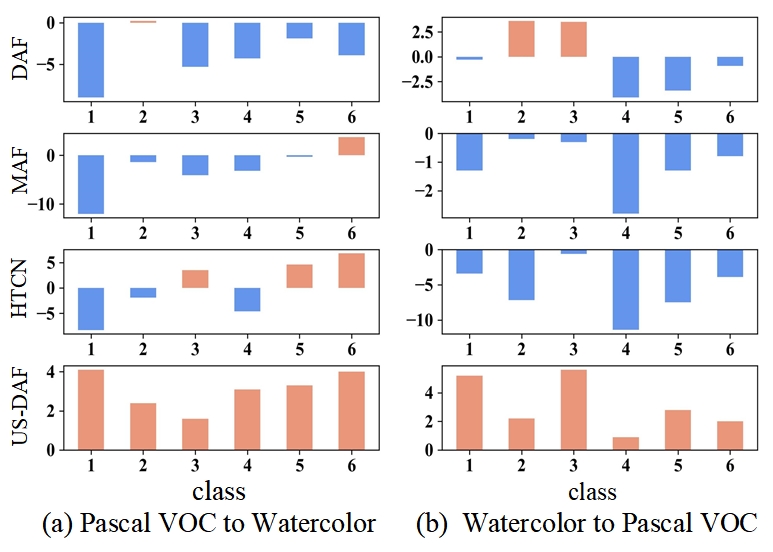}
  \caption{Algorithmic analysis. (a) Analysis of the negative transfer in cross-domain task PASCAL VOC $\rightarrow$ Watercolor. (b) Analysis of the negative transfer in cross-domain task Watercolor $\rightarrow$ PASCAL VOC.}\label{Figure5}
\end{figure}

{\bf Negative Transfer.}
In the practical setting of UniDAOD, most existing methods perform similarly to or even worse than Faster-RCNN without any adaptation, indicating that existing methods are prone to negative
transfer in UniDAOD settings. For example, Figure \ref{Figure5} (a) and (b) show the per-class accuracy gain compared to Faster RCNN on the tasks Pascal VOC $\rightarrow$ Waterclolor and Waterclolor $\rightarrow$ Pascal VOC. We can find that DAF, MAF, and HTCN suffer from negative transfer in most classes. Only US-DAF promotes positive transfer for all classes. This suggests that our proposed US-DAF has the capacity to quantify the class importance and intensify the common label set across domains.
\begin{figure*}[t]
  \centering
   \includegraphics[width=0.95\linewidth]{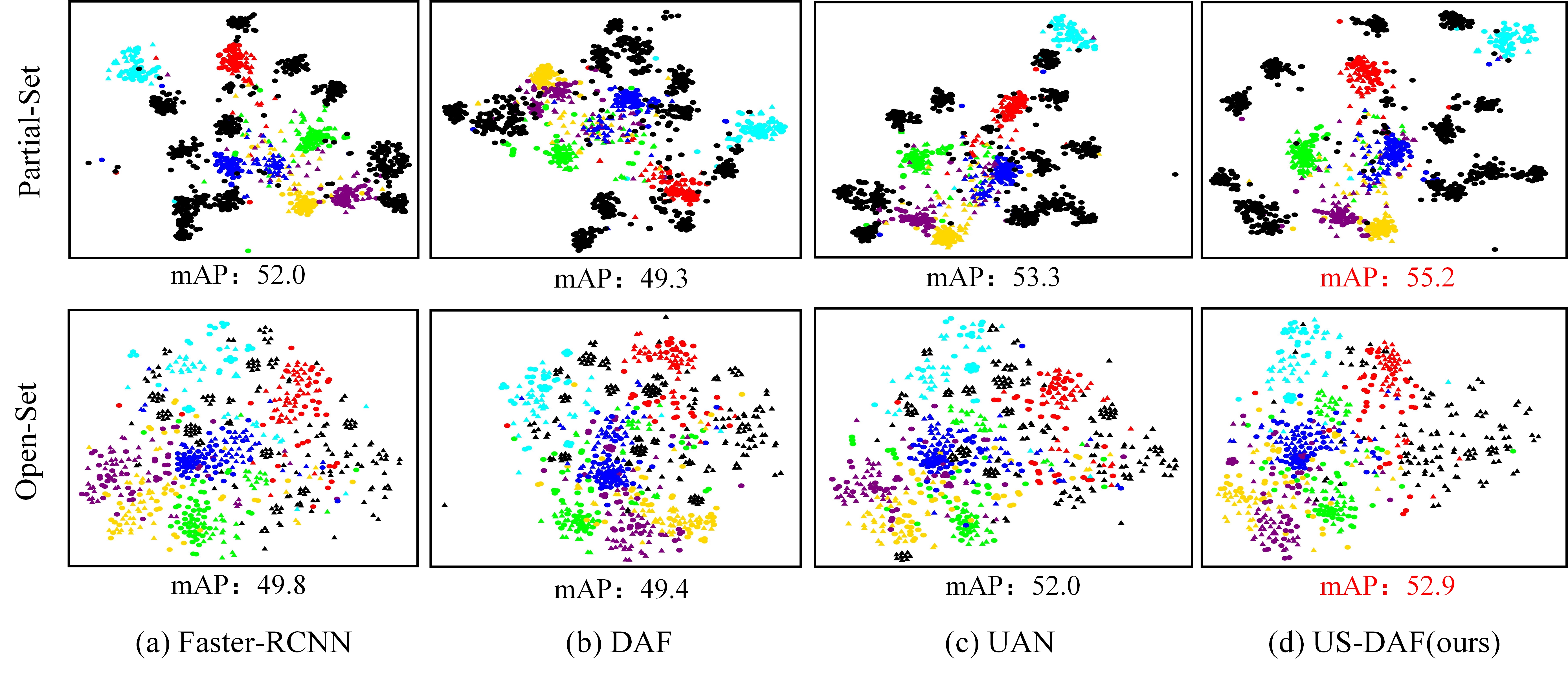}
   \setlength{\abovecaptionskip}{0.1cm}
   \caption{The t-SNE visualization on the Patrial-set and Open-set scenarios, where the circular ``$\circ$''  represent source samples  and the triangular ``$\vartriangle$'' represent target samples. Black dots are private-class samples while other colors are common-class samples. }
   \label{Figure6}
\end{figure*}

{\bf Visualization of Feature Distribution.} In Figure \ref{Figure6}, we used t-SNE \cite{ref42} to compare the distribution of induced features between our US-DAF and other models on the Patrial-set (\emph{i.e.}, Pascal VOC to Watercolor) and Open-set (\emph{i.e.}, Watercolor to Pascal VOC) scenarios, where different color stands for different common categories and the black dots stand for the private categories. We can observe that features of private classes and several common classes are close or even mixed together, indicating that DAF and UAN cannot discriminate known (common) and unknown (private) classes during training. By contrast, our proposed US-DAF produces features that can well separate the common and private classes, which benefits from the proposed strategy of filter mechanism and multi-label scale-aware adaptation.

\begin{figure}[t]
  \centering
   \includegraphics[width=0.85\linewidth]{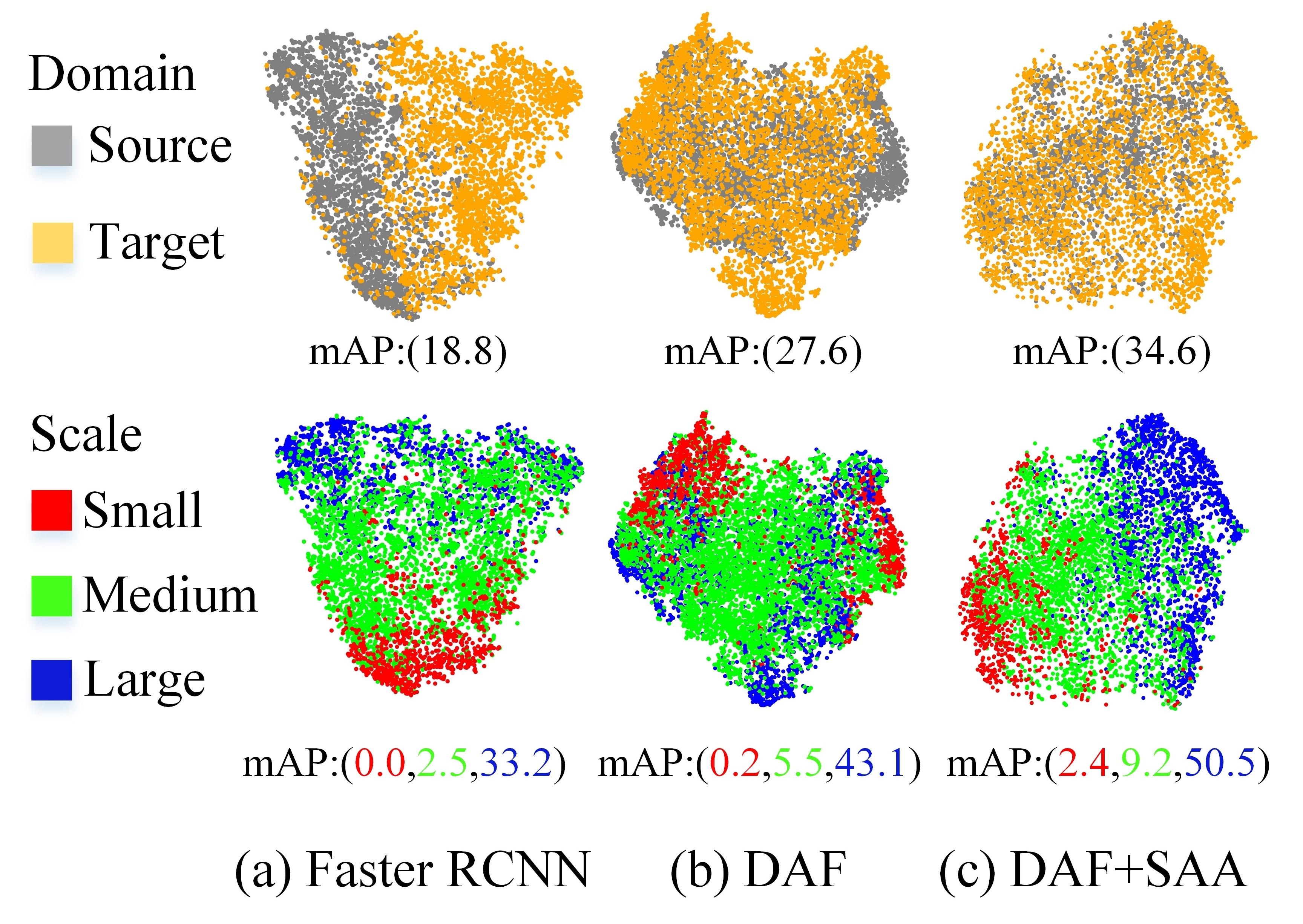}
   \setlength{\abovecaptionskip}{0.1cm}
   \caption{We show the features extracted with different models. The first row presents the domain-wise alignment, while the second row shows the scale-wise alignment. }
   \label{Figure7}
\end{figure}
{\bf Scale-Wise Analysis.}
In order to gain further insight into the influence of feature alignment, we conduct a scale-wise analysis, where we visualize the instance-level feature from different object
scales,  as well as provide a scale-wise quantitative evaluation.  As shown in Figure \ref{Figure7}, we use Cityscapes $\rightarrow$ Foggy Cityscapes in this study. On the top row  we show the domain-wise alignment. And on bottom row we show the scale-wise alignment, where we divide each instance into three sub-categories, based on the instance size: small ($\textless 20^2$ pixels), medium ($20^2 \sim 100^2$ pixels), large ($\textgreater 100^2$ pixels).

Figure \ref{Figure7} presents the results of the source-only Faster RCNN, DAF, and DAF+SAA.  From the results, the source and target features extracted from the source-only Faster R-CNN model can be clearly divided into two parts, and features of different scales are spanned across the feature space.   DAF performs a uniform domain alignment, which is agnostic to the scale. As a result, the features are aligned between the two domains to some extent. However, the alignment produces a side effect of wrongly aligned features across different scales. In contrast, our DAF+SAA  is able to take advantage of the scale information, and maintains the scale discriminability when aligning the features. This has resulted an observable better feature alignment. Further, We report mAP for each scale and summarize the results in Figure \ref{Figure7}. We observe that the proposed modules also demonstrate better quantitative results across scales.

\section{Conclusion}
In this paper, we introduce a novel setting that better meets the needs of real-world scenarios, Universal Domain Adaptive Object Detection (UniDAOD), which requires no prior knowledge on the label set of target domains. In order to meet this challenge of UniDAOD, we contribute a Universal Scale-Aware Domain Adaptive Faster R-CNN with Multi-Label Learning (US-DAF) framework,  which, to the best of our knowledge, is a pioneer work for object detection under both category shift and scale issue toward universal scenarios. In order to overcome the category shift of conventional UniDAOD, we introduce the filter mechanism to reject the private classes and preserve the common classes. Moreover, the scale-aware adapter is proposed with multi-label learning mechanism to tackle the problem caused by the large variety of scales in natural scenes. Through extensive experiments, we validated the effectiveness of our method by achieving a new state-of-the-art performance in various universal domain adaptation scenarios.

{\bf Acknowledgement.}
This work was partially supported by National Key R\&D Program of China (2021YFB3100800), Chongqing Natural Science Fund (cstc2021jcyj-jqX0023), CCF Hikvision Open Fund (CCF-HIKVISION OF 20210002), CAAI-Huawei MindSpore Open Fund, and Beijing Academy of Artificial Intelligence (BAAI).

{\small
\bibliographystyle{ieee_fullname}
\bibliography{egbib}
}

\end{document}